\documentclass[sigconf]{aamas}  

\renewcommand\footnotetextcopyrightpermission[1]{} 
\usepackage{graphicx}
\usepackage{subcaption}
\usepackage{enumerate}

\usepackage{amssymb}
\usepackage{amsmath}
\usepackage[makeroom]{cancel}

\usepackage{booktabs}
\usepackage{multirow}

\usepackage{xcolor}

\usepackage{algorithm}
\usepackage{algcompatible}

\usepackage{xspace}
\usepackage{cleveref}

\usepackage{todonotes}

\presetkeys%
    {todonotes}%
    {inline}{}

\newcommand{\rl}{Reinforcement Learning}
\newcommand{\brl}{Bayesian \rl}
\newcommand{\drl}{Deep \rl}
\newcommand{\mdp}{MDP}

\newcommand{\pomdp}{POMDP}

\newcommand{\fpomdp}{F-POMDP}
\newcommand{\ba}{Bayes-Adaptive}
\newcommand{\bapomdp}{BA-POMDP}

\newcommand{\fbapomdp}{FBA-POMDP}

\newcommand{\mc}{Monte-Carlo}

\newcommand{\dir}{Dirichlet}
\newcommand{\cpt}{CPT}

\newcommand{\bd}{BD}

\newcommand{\rs}{Rejection Sampling}
\newcommand{\is}{Importance Sampling}
\newcommand{\mh}{Metropolis-Hastings}
\newcommand{\gibbs}{Gibbs}
\newcommand{\mhgibbs}{MH-within-\gibbs}
\newcommand{\mcts}{\mc\ Tree Search}
\newcommand{\pomcp}{POMCP}
\newcommand{\bapomcp}{BA-POMCP}

\newcommand{\fbapomcp}{FBA-POMCP}

\newcommand{\tiger}{Tiger}
\newcommand{\ftiger}{Factored Tiger}
\newcommand{\ca}{Collision Avoidance}
\newcommand{\gw}{Gridworld}

\newcommand{\pomdpDef}{$\langle S, A, \Omega, D, R, \gamma, h \rangle$}
\newcommand{\fbapomdpDef}{$\langle \bar{S}, A, \Omega, \bar{D}, \bar{R}, \gamma, h \rangle$}

\newcommand{\ourmethod}{our method}
\newcommand{\bapomcpAgent}{\bapomcp}
\newcommand{\noreinvigAgent}{no-reinvigoration}
\newcommand{\cheatAgent}{knows-structure}

\providecommand{\funcName}[1]{\textsc{#1}}
\providecommand{\commentSymb}{//}

\newcommand{\hyperstate}{\langle s, G, \chi \rangle}
\newcommand{\newhyperstate}{\langle s', G', \chi' \rangle}
\newcommand{\history}{\langle \vec{a},\vec{o} \rangle}
\newcommand{\History}{\langle \vec{s}, \vec{a},\vec{o} \rangle}

\renewcommand{\todo}[1]{}

\title{\brl\ in Factored \pomdp s}
\author{Sammie Katt}
\affiliation{\institution{Northeastern University}}
\author{Frans Oliehoek}
\affiliation{\institution{Delft University of Technology}}
\author{Christopher Amato}
\affiliation{\institution{Northeastern University}}

\begin{document}

    \maketitle

    \section*{Abstract}
    Bayesian approaches provide a principled solution to the
exploration-exploitation trade-off in Reinforcement Learning. Typical
approaches, however, either assume a fully observable environment or
scale poorly. This work introduces the Factored \ba\ \pomdp\ model, a
framework that is able to exploit the underlying structure while
learning the dynamics in partially observable systems. We also present
a belief tracking method to approximate the joint posterior over state
and model variables, and an adaptation of the Monte-Carlo Tree Search
solution method, which together are capable of solving the underlying
problem near-optimally. Our method is able to learn efficiently given a
known factorization or also learn the factorization and the model
parameters at the same time. We demonstrate that this approach is able
to outperform current methods and tackle problems that were previously
infeasible.


    \todo{%
        TODO: Separate modelling (factored vs flat) from belief tracking
        (importance sampling, reinvigoration) and the planning part (MCTS,
        trivial look-ahead etc.). Maybe add a table where their combinations
        can be named. This can be reflect in the background where the
        separation could be emphasized, in the contribution section where we
        explicitly mention where our contributions fit, and in the experimental
        section where we should more clearly describe what the various methods
        do
    }

    \section{Introduction}
Robust decision-making agents in any non-trivial system must reason over
uncertainty in various dimensions such as action outcomes, the agent's
current state and the dynamics of the environment. The outcome and state
uncertainty are elegantly captured by \pomdp s~\cite{pomdp}, which enable
reasoning in stochastic, partially observable environments. \\

\todo{S\@: is the motivation for BRL  clear enough? Maybe consider: sample efficiency compared to DRL, usage of prior knowledge and exploration}

The \pomdp\ solution methods, however, assumes complete knowledge to the system
dynamics, which unfortunately are often not easily available. When such a model
is not available, the problem turns into a \rl\ (RL) task, where one must
consider both the potential benefit of learning as well as that of exploiting
current knowledge. Bayesian RL addresses this exploration-exploitation
trade-off problem in a principled way by explicitly considering the uncertainty
over the unknown parameters. While Model-based Bayesian RL have been applied to
partially observable models~\cite{ross2011bayesian}, these approaches do not
scale to problems with more than a handful of unknown parameters. Crucially,
they model the dynamics of the environment in a tabular fashion which are
unable to generalize over similar states and thus unable to exploit the
structure of real-world applications. 
Earlier work for \emph{fully observable} environments tackle this issue by
representing states with features and the dynamics as
graphs~\cite{structuredMDP}. Their formulation for the \mdp\ case, however,
does not accommodate for environments that are either partially hidden or where
the perception of the state is noisy. \\

\todo{S\@: is the the significance of our contribution clear enough?}

In this work we introduce the Factored \ba\ \pomdp\ (\fbapomdp), which captures
partially observable environments with unknown dynamics, and does so by
exploiting structure. Additionally we describe a solution method based on the
\mcts\ family and a mechanism for maintaining a belief specifically for the
\fbapomdp. We show the favourable theoretical guarantees of this approach and
demonstrate empirically that it outperforms the current state-of-the-art
methods. In particular, our method outperforms previous work on $3$ domains, of
which one is too large to be tackled by solution methods based on the tabular
\bapomdp.

    \section{Background}
We first provide a summary of the background literature. This section is
divided into an introduction to the \pomdp\ and \bapomdp, typical solution
methods for those models, and factored models.

\subsection{The \pomdp\ \& \bapomdp}

The \pomdp~\cite{pomdp} is a general model for decision-making in
stochastic and partially observable domains, with execution unfolding over
(discrete) time steps. At each step the agent selects an action that
triggers a state transition in the system, which generates some reward and
observation. The observation is perceived by the agent and the next time
step commences.
Formally, a \pomdp\ is described by the tuple \pomdpDef, where
    $S$ is the set of states of the environment; $A$ is the set of actions;
    $\Omega$ is the set of observations;
    $D$ is the `dynamics function' that describes the dynamics of the
    system in the form of transition probabilities $D(s',o|s,a)$;\footnote%
    {%
        This formulation generalizes the typical formulation with separate
        transition $T$ and observation functions $O$: $D = \langle T, O
        \rangle$. In our experiments, we do employ this typical
        factorization.
    }
    $R$ is the immediate reward function $R(s, a)$ that describes the
    reward of selecting $a$ in $s$;
    $\gamma \in (0,1)$ is the discount factor; and 
    $h$ is the horizon of an episode in the system. 
In this description of the \pomdp, $D$ captures the probability of
transitioning from state $s$ to the next state $s'$ and generating
observation $o$ in the process for each action $a$.\\

\todo{overall perhaps change / improve the notation for history ($\history_0^t$)}

The goal of the agent in a \pomdp\ is to maximize the expectation over the
cumulative (discounted) reward, also called the return. The agent has no
direct access to the system's state, so it can only rely on the
\emph{action-observation history} $h_t=\history_0^t$ up to the current step
$t$. It can use this history to maintain a probability distribution over
the state, also called a belief, $b(s)$. 
A solution to a \pomdp\ is then a mapping from a belief $b$ to an action
$a$, which is called a $policy$ $\pi = p(a|b)$. Solution methods aim to
find an optimal policy, a mapping from a belief to an action with the
highest possible expected return. \\

The \pomdp\ allows solution methods to compute the optimal policy given a
complete description of the dynamics of the domain. In many real world
applications such a description is not readily available. The
\bapomdp~\cite{ross2011bayesian} is a model-based \brl\ framework to model
applications where those are hard to get, allowing the agent to directly
reason about its uncertain over the \pomdp\ model.
Conceptually, if one observed both the states and observations, then we
could count the number of the occurrences of all $\langle s, a, s', o
\rangle $ transitions and store those in $\chi$, where we write
$\chi_{sa}^{s'o}$ for the number of times that $s,a$ is followed by $s',o$.
The belief over $D$ could then compactly be described by \dir\
distributions, supported by the counts $\chi$. While the agent cannot
observe the states and thus has uncertainty about the actual count vector,
this uncertainty can be represented using regular \pomdp\ formalism. That
is, the count vector is included as part of the hidden state of the \pomdp.
\\

Formally, the \bapomdp\ is the \pomdp\ \fbapomdpDef\ with (hyper-) state
space $\bar{S} = \langle S, X \rangle$, where $X$ is the countably infinite
space of assignments of $\chi$. While the observation and action space
remain unchanged, a state in the \bapomdp\ now includes \dir\ parameters:
$\bar{s} = \langle s, \chi \rangle$. The reward function still only depends
on the underlying \pomdp\ state: $\bar{R}(\bar{s},a) = R(s,a)$. The
dynamics of the \bapomdp, $\bar{D}=p(s',\chi,o|s,\chi,a)$, factorize to
$p(s',o|s,\chi,a)p(\chi'|s,\chi,a,s',o)$, where $p(s',o|s,\chi,a)$
corresponds to the expectation of $D_{sa}^{s'o}$ according to $\chi$:

\begin{equation}\label{eq:bapomdp-domain-dynamics}
    p(s',o|s,\chi,a) = P_\chi(s',o|s,a) = \frac{\chi_{sa}^{s'o}}{\sum_{s'o}\chi_{sa}^{s'o}}
\end{equation}

If we let $\delta_{sa}^{s'o}$ denote a vector of the length of $\chi$
containing all zeros except for the position corresponding to $\langle
s,a,s',o \rangle$ (where it is $1$), and if we let $\mathbb{I}_{a}(b)$
denote the Kronecker delta that indicates (is $1$ when) $a=b$, then we
denote $\mathcal{U}(\chi,s,a,s',o) = \chi+\delta_{sa}^{s'o}$ and can write
$p(\chi|s,\chi,a,s',o)$ as
$\mathbb{I}_{\chi'}(\mathcal{U}(\chi,s,a,s',o))$. Lastly, just like any
Bayesian method, the \bapomdp\ requires a prior $\bar{b}_0$, the initial
joint belief over the domain state and dynamics. Typically the prior
information about $D$ can be described with a single set of counts
$\chi_0$, and $\bar{b}_0$ reduces to the joint distribution
$b_0(s)\times\chi_0$ where $b_0(s)$ is the distribution over the initial
state of the underlying \pomdp.

\subsection{Learning by Planning in \bapomdp s}

The countably infinite state space of the \ba\ model poses a challenge to
offline solution methods due to the curse of dimensionality. Partially
Observable Monte-Carlo Planning (\pomcp)\cite{POMCP}, a Monte-Carlo Tree
Search (MCTS) based algorithm, does not suffer from this curse as its
complexity is independent of the state space. As a result, the extension to
the Bayes-Adaptive case, \bapomcp~\cite{katt2017learning}, has shown
promising results. \\

\pomcp\ incrementally constructs a look-ahead action-observation tree using
\mc\ simulations of the POMDP\@. The nodes in this tree contain statistics
such as the number of times each node has been visited and the average
(discounted) return that follows. Each simulation starts by sampling a
state from the belief, and traverses the tree by picking an actions
according to UCB~\cite{auer2002finite} and simulating observations
according to the POMDP model. Upon reaching a leaf-node, the tree is
extended with a node for that particular history and generates an estimate
of the expected utility of the node. The algorithm then propagates the
accumulated reward back up into the tree and updates the statistics in each
visited node. The action selection terminates by picking the action at the
root of the tree that has the highest average return. \\

The key modifications of the application \pomcp\ to \bapomdp s are
two-fold: (1) a simulation starts by sampling a hyper-state $\langle s,\chi
\rangle$ at the start and (2) the simulated step follows the dynamics of
the \bapomdp\ (\cref{alg:bapomcp-step}). During this step first the domain
state transitions and an observation is generated according to $\chi$
(\cref{alg:bapomcp-step} line~\ref{alg-line:bapomcp-domain-step}), which in
turn are then used to update the counts (\cref{alg:bapomcp-step}
line~\ref{alg-line:bapomcp-counts-step}).

\todo{S\@: We do not mention or use root sampling here, mainly because it is
sort of orthogonal to this work. Should we do that anyway, at the expense of
brevity?}

\begin{algorithm}[ht!]
    \caption{\funcName{\bapomcp-step}}\label{alg:bapomcp-step}
    \begin{algorithmic}[1]
        \Statex{\textbf{Input} $s$: domain state, $\chi$: \dir s over $D$}
        \Statex{\textbf{Input} $a$: simulated action}
        \Statex{\textbf{Output} $s'$: new domain state, $\chi'$: updated $\chi$}
        \Statex{\textbf{Output} $o$: simulated observation}
        \STATE{\commentSymb\ \cref{eq:bapomdp-domain-dynamics}}
        \STATE{$s',o' \sim P_{\chi}(\cdot|s,a)$}\label{alg-line:bapomcp-domain-step}
        \STATE{$\chi' \gets \chi + \delta_{sa}^{s'o}$}\label{alg-line:bapomcp-counts-step}
        \STATE{\textbf{return} $s',\chi',o$}
    \end{algorithmic}
\end{algorithm} 

\todo{TODO: make comments on \is\ inline, removing pseudocode}

Given enough simulations, \bapomcp\ converges to the optimal solution with
respect to the belief it is sampling states from~\cite{katt2017learning}. One
can compute this belief naively in closed form in finite state spaces by
iterating over all the possible next states using the model's
dynamics~\cite{ross2011bayesian}. This quickly becomes infeasible and is only
practical for very small environments. More common approaches approximate the
belief with \emph{particle filters}~\cite{Thrun99}. There are numerous methods
to update the particle filter after executing action $a$ and receiving
observation $o$, of which \emph{\rs} has traditionally been used for
(BA-)POMCP\@. \emph{\is}~\cite{gordon1993novel} (outlined
in~\cref{alg:importance-sampling}), however, has been shown to be superior in
terms of the chi-squared distance~\cite{chen2005another}. \\

\begin{algorithm}[ht!]
    \caption{\funcName{\is}}\label{alg:importance-sampling}
    \begin{algorithmic}[1]
        \Statex{\textbf{Input} $K$: number of particles, $\bar{b}$: current belief}
        \Statex{\textbf{Input} $a$: taken action, $o$: real observation}
        \Statex{\textbf{Output} $\bar{b}$: updated belief, $\mathcal{L}$: update likelihood}
        \STATE{$\bar{b}' \gets \{\}$}
        \STATE{$\mathcal{L} \gets 0$}
        \STATE{\commentSymb\ update belief}\label{alg-line:update}
        \FOR{$\langle \bar{s},w \rangle \in \bar{b}(\bar{s})$}
            \STATE{$ \bar{s}'  \sim p_{\bar{D}}(\cdot|\bar{s},a)$}
            \STATE{$w' \gets p_{\bar{D}}(o|\bar{s},a,\bar{s}') w$}\label{alg-line:is-weighting}
            \STATE{$\mathcal{L} \gets \mathcal{L} + w'$}
            \STATE{add $\langle \bar{s}, w' \rangle$ to $\bar{b}'$}
        \ENDFOR{}
        \STATE{\commentSymb\ resample step}\label{alg-line:resample}
        \STATE{$\bar{b} \gets \{\}$}
        \FOR{$i \in 1 \dots K$}
            \STATE{$\bar{s} \sim \bar{b}'(\bar{s}$)}
            \STATE{$add$ $\langle \bar{s}, w{=}\frac{1}{K} \rangle$ to $\bar{b}$}
        \ENDFOR{}
        \STATE{$\textbf{return}$ $\bar{b}, \mathcal{L}$}
    \end{algorithmic}
\end{algorithm}

\todo{S\@ @ F\@: I know this is very detailed, but we refer back to it
(specifically $\mathcal{L}$) during the explanation of our MH-Gibbs update
formulae. Does that justify it a little?}

In \is\ the belief is represented by a weighted particle filter, where each
particle $x$ is associated with a weight $w_x$ that represents its probability
$p(x) {=} \frac{w_x}{\sum_{i=1}^{K}w_i}$. \is\ re-computes the new belief given
an action and observation with respect to the model's dynamics
$p_{\bar{D}}(\bar{b}'|\bar{b},a,o)$ in two steps. First, each particle is
updated using the transition dynamics $p_{\bar{D}}(\bar{s}'|\bar{s},a)$, and
then weighted according to the observation dynamics
$p_{\bar{D}}(o|\bar{s},a,\bar{s}')$. Note that the sum of weights  of the
belief after this step $\mathcal{L}^t {=} \sum w_i^t$ represents the likelihood
of the belief update at time $t$. The likelihood of the entire belief given the
observed history can be seen as the product of the likelihood of each update
step $\mathcal{L}_{h^t} = \mathcal{L}^t \mathcal{L}_{h^{t-1}}$. In the second
step, starting on~\cref{alg-line:resample} of~\cref{alg:importance-sampling},
the belief is resampled, as is the norm in sequential \is.

\subsection{Factored Models}

Just like most multivariate processes, the dynamics of the \pomdp\ can
often be represented more compactly with graphical models than by tables:
conditional independence between variables leads to the reduction of the
parameter space, leading to simpler and more efficient models. 
The Factored \pomdp\ (\fpomdp)~\cite{boutilier1996computing} represents the
states and observations with features and the dynamics $D$ as a collection
of Bayes-Nets (BN), one for each action. \\

Let us denote the featured state space $S = \{S^1, \dots, S^n\}$ into $n$
features, and observation space $\Omega = \{\Omega^1, \dots, \Omega^m \}$
into $m$ features. Then, more formally, a BN as a dynamics model for a
particular action consists of an input node for each state feature $s^i$
and an output node for each state and observation feature $s'$ and $o$. The
topology $G \in \mathcal{G}$ describes the directed edges between the
nodes, of which the possible graphs in is restricted such that the input
nodes $s$ only have outgoing edges and the observation nodes $o$ only have
incoming edges. For simplicity reasons we also assume that the output state
nodes $s'$ are independent of themselves. The \emph{Conditional Probability
Tables} (\cpt s) $\theta$ describe the probability distribution over the
values of the nodes given their (input) parent values $PV_G()$. The
dynamics of a \fpomdp\ are then defined as follows

$$D(s',o|s,a) = [\prod\limits_{s_i' \in s'}P_{\theta^a}(s_i' | PV_{G^a}(s))]
[\prod\limits_{o_i \in o}P_{\theta^a}(o_i | PV_{G^a}(s'))],$$

Some approaches are able to exploit the factorization of \fpomdp s, which
typically leads to better solution~\cite{boutilier1996computing}. These
methods, however, operate under the assumption that the dynamics are known
a-priori and hence cannot be applied to applications where this is not the
case.

    \section{Bayesian RL in Factored \pomdp s}
The \bapomdp\ provides a Bayesian framework for RL in \pomdp s, but is
unable to describe or exploit structure that many real world applications
exhibit. It also scales poorly, as the number of parameters grow quadratic
in the state space, $O(|S|^2|A||\Omega|)$, where only one parameter (count)
is updated after each observation. Here we introduce the Factored \bapomdp\
(\fbapomdp), the \ba\ framework for the factored \pomdp, that is able to
model, learn and exploit such relations in the environment. 

\subsection{The Factored \bapomdp}

If we consider the case where the structure $G$ of the underlying \pomdp\
is known a-priori, but its parameters $\theta$ are not, then it is clear
that we could define a Bayes-Adaptive model where the counts describe \dir\
distributions over the \cpt s: $\chi_G$ (which we know how to maintain
over time). However, this assumption is unrealistic, so we must also
consider both the topology $G$ and its parameters $\chi$ as part of the
hidden parameters, in addition to the domain state $s$. \\

\todo{make notation and use of $D$ consistent with background}

We define the \fbapomdp\ as a \pomdp\ with the (hyper-) state space
$\bar{S} = S \times \mathcal{G} \times X$. Let us first consider its
dynamics $\bar{D} = p(\newhyperstate,o | \hyperstate, a)$. This joint
probability can be factored using the same standard independence
assumptions made in the \bapomdp:

\begin{align}
    \bar{D}(\bar{s}'{=}\newhyperstate,o|\bar{s},a) &= \nonumber \\
    & p(s',o|\hyperstate,a)  \label{eq:fbapomdp-D} \\
    & p(\chi'|\hyperstate,a,s',G',o)  \label{eq:fbapomdp-chi} \\
    & p(G'|\hyperstate,a,s',o) \label{eq:fbapomdp-G}
\end{align}

Term $p(s',o|\dots)$ (\cref{eq:fbapomdp-D}) corresponds to the expectation
of $p(s',o|s)$, under the joint \dir\ posterior $\chi_{G^a}$ over \cpt s
$\theta_{G^a}$. Given the expected \cpt s $\mathbf{E}(\chi_{G^a}) =
\hat{\theta}_{G^a}$, that probability is described by the product governed
by the topology: $P_{\hat{\theta}_{G^a}}(s',o|s) = \prod\limits_{x \in
s',o} \hat{\theta}_{G^a}^{x|PV(x)}$. Throughout the rest of the paper we
will refer to this probability~\cref{eq:fbapomdp-D} with $D_{\chi_G}$. \\

$p(\chi'|\dots)$ (\cref{eq:fbapomdp-chi}) describes the update
operation on the counts $\chi$ that correspond to $\langle s,a,s',o
\rangle$: $\mathcal{U}(\chi_{G},s,a,s',o)$. Note that this will update
$n+m$ counts, one for each feature. Lastly, we assume the topology of $G$
is static over time, which reduces $p(G'|\dots)$ (\cref{eq:fbapomdp-G}) to
the Kronecker delta function $\mathbb{I}_{G'}(G)$. This leads to the
following definition of the \fbapomdp\ model, given tuple \fbapomdpDef:

\begin{itemize}
    \item 

        $A$, $\gamma$, $h$:
        Identical to the underlying \pomdp.

    \item $\bar{R}(\bar{s},a) = R(s,a)$ ignores the counts and reduces to
        the reward function of the \pomdp\ just like in the \bapomdp.

    \item 

        $\bar{\Omega}$: $\{ \Omega^0 \times \cdots \times \Omega^m \}$. Set
        of possible observations defined by their features.

    \item 

        $\bar{S}$: $\{ S^0 \times \cdots \times S^n \} \times \mathcal{G}^A
        \times X_{G^A}$. The cross product of the domain's factored state
        space and the set of possible topologies, one for each action $a$,
        and their respective \dir\ distribution counts.

    \item

        $\bar{D}$: $p(\bar{s}',o|\bar{s},a) = D_{\chi_G}(s,a,s',o)
        \mathbb{I}_{\chi_G'}(\mathcal{U}(\chi_{G},s,a,s',o))
        \mathbb{I}_{G'}(G)$, as described above.

\end{itemize}

A prior for the \fbapomdp\ is a joint distribution over the hyper-state
$\bar{b}_0(\hyperstate)$. In many applications the influence relationships
between features is known a-priori for large parts of the domain. For the
unknown parts, one could consider a uniform distribution, or distributions
that favours few edges.

\subsection{Solving \fbapomdp s}

Solution methods for the \fbapomdp\ face similar challenges as those for
\bapomdp s with respect to uncountable large (hyper-) state spaces as a
result of the uncertainty over current state and the dynamics. So it is
only natural to turn to \pomcp-based algorithms for inspiration. \\

\bapomcp\ extends MCTS to the Bayes-Adaptive case by initiating simulations
with a $\langle s,\chi \rangle$ sample (from the belief) and applying the
\bapomdp\ dynamics to govern the transitions (recall~\cref{alg:bapomcp-step}).
We propose a similar \pomcp\ extension, the \fbapomcp, for the factored
case where we sample a hyper-state $\hyperstate $ at the start of each
simulation, and apply the \fbapomdp\ dynamics $\bar{D}$ to simulate steps.
This is best illustrated in~\cref{alg:fba-pomcp}, which replaces
\funcName{\bapomcp-step}~\cref{alg:bapomcp-step}. During a step the sampled
$\langle G,\chi \rangle$ is used to sample a transition, after which the
counts associated with that transition are updated.

\begin{algorithm}[!ht]
    \caption{%
        \funcName{\fbapomcp-step}
    }\label{alg:fba-pomcp}
        \begin{algorithmic}[1]
            \Statex{\textbf{Input} $s$: domain state, $G$: graph topology}
            \Statex{\textbf{Input} $\chi$: \dir s over \cpt s in $G$, $a$: simulated action}
            \Statex{\textbf{Output} $s'$: new domain state, $G'$: $G$}
            \Statex{\textbf{Output} $\chi'$: updated $\chi$, $o$: simulated observation}
            \STATE{$s',o' \sim p_{\bar{D}^{\chi_G}}(\cdot|s,a)$}
            \STATE{\commentSymb\ increment count of each node-parent in place}
            \FOR{each node $n \in s',o$ and its value $v$}
            \STATE{$\chi_{G^a}^{n,v|PV(n)} \gets \chi_{G^a}^{n,v|PV(n)}+1 $} 
            \ENDFOR\
            \STATE{$s \gets s' $}
            \STATE{\textbf{return} $\hyperstate,o$}
        \end{algorithmic}
\end{algorithm}

\todo{mention that linking states, root sampling \& expected also applies}

\subsection{Belief tracking \& Particle Reinvigoration}

Because structures in the particles are not updated over time and due to
particle degeneracy, traditional particle filter belief update schemes then to
converge to a single structure, which is inconsistent with the true posterior,
leading to poor performance.
To tackle this issue, we propose a MCMC-based sampling scheme to occasionally
reinvigorate the belief with new particles according to the (observed) history
$p(\hyperstate| \history,\bar{b}_0)$. \\

First we introduction the notation $\vec{x}_{r \dots t}$ which describes
the (sequence of) values of $x$ from time step $r$ to $t$ of real
interactions with the environment, with the special case of $x_t$, which
corresponds to the value of $x$ at time step $t$ (where $\vec{x}$ can be a
sequence of states, action or observations). For brevity we also use
`model' and the tuple $\langle G,\chi \rangle$ interchangeably in this
section, as they represent the dynamics of a \pomdp. Lastly, we refer to
$T$ as the last time step in our history. \\

On the highest level we apply \gibbs\ sampling, which approximates a joint
distribution by sampling variables from their conditional distribution with the
remaining variables fixed: we can sample $p(x,y)$ by picking some initial $x$,
and sampling $y \sim p(y|x)$ followed by $x \sim p(x|y)$. Here we pick
$x{=}\vec{s}$ and $y{=}\langle G,\chi \rangle$ and sample alternatively a model
given a state sequence and a state sequence given a model:

\begin{enumerate}[i.]
    \item $\vec{s} \sim p(\cdot | G, \chi, \history, \bar{b}_0)$\label{item:s-distribution}
    \item $G,\chi \sim p(\cdot| \History, \bar{b}_0)$\label{item:model-distribution}
\end{enumerate}

\textbf{Sample step (\ref{item:s-distribution})} samples a state-sequence given
the observed history $\history$ and current model. A very simple and naive
approach is to use \rs\ to sample state and observation sequences based on the
action history and reject them based on observation equivalence, optionally
exploiting the independence between episodes. Due to the high rejection rate,
this approach is impractical for non-trivial domains. Alternatively, we model
this task as sampling from a Hidden Markov Model, where the transition
probabilities are determined by the model $\langle G,\chi \rangle$ and action
history $\vec{a}$. \\

To sample a hidden state sequence from an HMM given some observations (also
called \emph{smoothing}) one typically use forward-backward messages to compute
the conditionals probability of the hidden states
efficiently~\cite{rabiner1986introduction}. We first compute the conditional
$\forall t: p(s_t | \vec{a}_{t \dots T}, \vec{o}_{t \dots T},G,\chi)$ with
backward-messages and then sample $s_0 \dots s_T$ hierarchically in a single
forward pass. \\

\todo{describe messages}

\textbf{The second conditional (\ref{item:model-distribution})} in the \gibbs\
sampling scheme is from distribution $p(G,\chi| \History, \bar{b}_0)$, which
itself is split into two  sample steps. We first \textbf{(a)} sample ${G} \sim
p(\cdot | \History, \bar{b}_0)$ using \mh. This is followed by sampling a set
of counts \textbf{(b)} $\chi \sim p(\cdot | \History, G, \bar{b}_0)$, which is
a deterministic function that simply takes the prior $\chi_0^G$ and counts the
transitions in the history $\History$. For the first sample step
\textbf{(\ref{item:model-distribution} a)} ${G} \sim p(\cdot | \History,
\bar{b}_0)$ we start from the general \mh\ case: \\

\mh\ samples some distribution $p(x)$ using a proposal distribution
$q(\tilde{x}|x)$ and testing operation. The acceptance test probability of
$\tilde{x}$ is defined as
$\frac{p(\tilde{x})q(\tilde{x}|x)}{p(x)q(x|\tilde{x})}$. More specifically,
given some initial value $x$, \mh\ consists of:

\begin{enumerate}[(1)]
    \item sample $\tilde{x} \sim q(\tilde{x}|x)$
    \item with probability $\frac{p(\tilde{x})q(\tilde{x}|x)}{p(x)q(x|\tilde{x})}$: $x \gets \tilde{x}$
    \item store $x$ and go to (1)
\end{enumerate}

Let us take $p(x)$ as $p(G|\History,\bar{b}_0)$ and $q$ to be domain
specific and symmetrical, then we derive the following \mh\ step for
\textbf{(\ref{item:model-distribution} a)}:

\begin{align*}
    MH\text{-}STEP_{accpt} &= \frac{p(\tilde{G}| \History, \bar{b}_0)\cancel{q(\tilde{x}|x)}}{p(G| \History, \bar{b}_0)\cancel{q(x|\tilde{x})}}  \\
    &= \frac{\frac{p(\History,\tilde{G}|\bar{b}_0)}{\cancel{p(\History|\bar{b}_0)}}}{\frac{p(\History,G|\bar{b}_0)}{\cancel{p(\History|\bar{b}_0)}}} \\
    &= \frac{p(\History,\tilde{G}|\bar{b}_0)}{p(\History,G|\bar{b}_0)}
\end{align*}

\todo{go in further detail on the BD-score, verify and show it represents the
likelihood} 

Which leads to the likelihood ratio between the two graph structures. It has
been shown that the likelihood $p(\History,G|\bar{b}_0)$, given some mild
assumptions (such as that the prior is a \dir), this value is given by the
\bd-score metric~\cite{heckerman1995learning}: $P(G|D) \propto P(G,D) =
BD(G,D)$. Given some initial set of prior counts for $G$, $\chi_0$, and a
dataset of occurrences $N^{nev}$ of values $v$ with parent values $e$ for node
$n$ provided by $\History$, then the score is computed as follows:
$p(\History,G|\bar{b}_0) = BD(G,\History|\bar{b}_0) = $

\begin{equation*}
    \prod\limits_{n}\prod\limits_{e}
    \frac{\Gamma(\chi^{ne}_0)}{\Gamma(\chi^{ne}_0+N^{ne})} \prod\limits_{v}
    \frac{\Gamma(\chi^{nev}_0+N^{nev})}{\Gamma(\chi^{nev}_0)}
\end{equation*}

Where we abuse notation and denote the total number of counts,
$\sum\limits_{v} \chi^{nev}$, as $\chi^{ne}$ (and similarly $N^{ne} =
\sum\limits_{v} N^{nev}$). \\

Given this acceptance probability, \mh\ can sample a new set of graph
structures $G$ with corresponding counts for the CPTs $\chi$.  This particular
combination of MCMC methods --- \mh\ in one of \gibbs's conditional sampling
steps --- is also referred to as \mhgibbs\ and, surprisingly, has shown to
converge to the true distribution even if the \mh\ part only consist of 1
sample per
step~\cite{lee2002particle,martino2015independent,tierney1994markov,robert2013monte,liang2011advanced}.
\\

The overall particle reinvigoration procedure, assuming some initial
$\langle G, \chi \rangle$, is as follows:

\begin{enumerate}
    \item sample from HMM $p(\vec{s}|\history,G,\chi,\bar{b}_0)$
    \item sample from MH\@: $p(G|\History,\bar{b}_0)$ (using BD-scores)
    \item compute counts: $p(\chi|\History,G,\bar{b}_0)$
    \item add $\hyperstate$ to belief and go to $1$
\end{enumerate}

It is not necessary to do this operation at every time step, instead the
$\log$-likelihood $\mathcal{L}$ of the current belief is a useful metric to
determine when to resample. Fortunately, it is a by-product of \is\ at
line~\ref{alg-line:is-weighting} of~\cref{alg:importance-sampling}. The
total accumulated weight, denoted as $\eta = \sum w^i$ (the normalization
constant) is the likelihood of the belief update. Starting with
$\mathcal{L}{=}0$ at $t{=}0$, we maintain the likelihood over time
$\mathcal{L} = \mathcal{L} + \log{\eta_t}$ and update the posterior
$b(\hyperstate|\history,\bar{b}_0)$ whenever the $\mathcal{L}$ drops below
some threshold.

\subsection{Theoretical guarantees}

Here we consider $2$ theoretical aspects of our proposed solution method.
We first note that \fbapomcp\ converges to the optimal solution with
respect to the belief, and secondly point out that the proposed belief
tracking scheme converges to the true belief. \\

Analysis from~\cite{POMCP} proof that the value function constructed by
\pomcp, given some suitable exploration constant $u$, converges to the
optimal value function with respect to the initial belief. Work on
\bapomcp~\cite{katt2017learning} extends the proofs to the \bapomdp.
Their proof relies on the fact that the \bapomdp\ is a \pomdp\ (that
ultimately can be seen as a belief \mdp), and that \bapomcp\ simulates
experiences with respect to the dynamics $\bar{D}$. These notions also
apply to \fbapomdp\ and we can directly extend the proofs to our solution
method. \\

Given that \fbapomcp\ converges to the optimal value function with respect
to the belief, it is important to consider whether that the belief as a
result of our particle reinvigoration approximates the true posterior (note
that we are only concerned with the reinvigoration part of the belief
update, as it is widely known that \is\ with particle filters is unbiased).
This follows directly from the convergence properties of \gibbs\ sampling,
\mh\ and \mhgibbs\ that have been used to sample from the posterior. Since
these methods are unbiased approximations and we use them directly to
sample from the true posterior $p(\hyperstate|\history,\bar{b}_0)$, we show
that our solution method converges to the true distribution (given the
initial belief).

    \section{Experiments}
Here we provide empirical support for our factored Bayes-Adaptive approach
on $3$ domains: the \ftiger, \ca, and \gw\ problem. The \ftiger\ problem is
an extension of the well-known \tiger\ problem~\cite{pomdp}, the \ca\
problem is taken from~\cite{luo2016importance} and the \gw\ is inspired by
navigational tasks. In this section we first describe each domain on a high
level, followed by the prior information we assume given to the agent. For
more details please refer to the appendix.

\todo{TODO: say something about parameters, also that we use root sampling}

\subsection{Setup}

The \tiger\ domain describes a scenario where the agent is faced with the
task of opening one out of two doors. Behind one door lurks a tiger, a
danger and reward of $-100$ that must be avoided, while the other door
opens up to a bag of gold for a reward of $10$. The agent can choose to
open either doors (which ends the episode) or to listen for a signal: a
noisy observation for a reward of $-1$. This observation informs the agent of
the location of the tiger with $85\%$ accuracy. In the \textbf{\ftiger}
domain we increase the state space artificially by adding $7$ uninformative
binary state features. While these features increase the state space, they
are stationary over time and do not affect the observation function. From a
planning point of view the problem retains its complexity regardless of the
number of features, the challenge for a learning agent however is to infer
the underlying dynamics in the significantly large domain. \\

In this particular case, the agent is unsure about the observation
function. In particular, the prior belief of the agent assigns $60\%$
probability to hearing the tiger correctly. The prior belief over the
structure of the observation model is uniform. This means that each edge
from any of the $8$ state features to the observation feature has a $50\%$
chance of being present in a particle in the initial belief. \\

In the \textbf{\ca} problem the agent pilots a plane that flies from
(centre figure) right to left ($1$ cell at a time) in a $2$-dimensional
grid. The agent can choose to stay level for no cost, or move either
diagonally up or down with a reward of $-1$. The episode ends when the
plane reaches the column on the left, where it must avoid collision with a
vertically moving obstacle (or face a reward of $-1000$). The obstacle
movement is stochastic, and the agent observes its location at each time
step with some noise. The optimal policy attempts to avoid the obstacle
with as little vertical movement as possible. \\

While we assume the agent knows the observation and transition model of the
plane, the agent initially underestimates the movement strategy of the
obstacle: it believes it will stay put $90\%$ of the time and move either
direction with $5\%$ probability each, while the actual probabilities are
respectively $50\%$ and $25\%$. The agent knows that the location of the
obstacle in the next state depends on its previous location a-priori, but
otherwise assume no initial acknowledge on the structure of the model with
respect to the location of the object. \\

\textbf{\gw}, is a $2$-dimensional grid in which the agent starts in the
bottom left corner and must navigate to a goal cell. The goal cell is
chosen from a set of candidates at the start of an episode, and can be
fully observed by the agent. The agent additionally observes its own
location with a noisy localizer. The agent can move in all $4$ directions,
which are generally successful $95\%$ of the attempts. There are, however,
specific cells that significantly decrease the chance of success to $15\%$,
essentially trapping the agent. The target of the agent is to reach the
goal as fast as possible. \\

In this domain we assume no prior knowledge of the location or the number
of `trap' cells and the prior assigns $95\%$ probability of transition
success on all cells. The observation model in this domain is considered
known. Here we factor the state space up into the location of the goal
state and the $(x,y)$ position of the agent and assume the agent knows that
its next location is dependent on the previous, but is unsure whether the
goal location is necessary to model its transition probabilities. This
results in a prior belief where all particles contain models where, for
each action, the $x$ and $y$ values of the current location are used to
predict the next location of the agent, and half the particles \emph{also}
include the value of the goal cell as input edge.

\todo{add baseline description and results}

\begin{figure*}[t!]
    \centering
    \begin{subfigure}[t]{0.32\textwidth}
        \centering
        \includegraphics[width=1.05\linewidth]{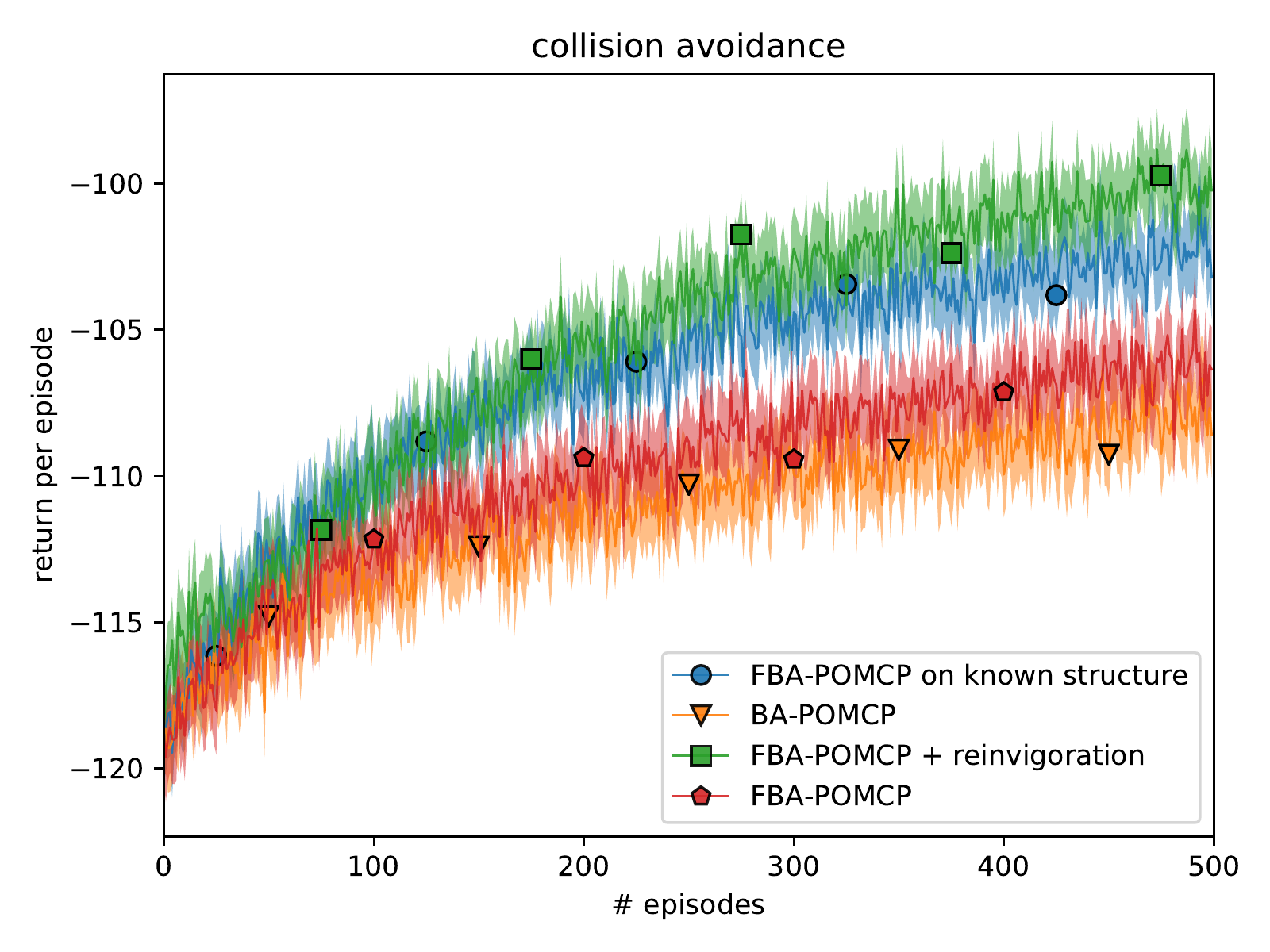}
    \end{subfigure}%
    ~ 
    \begin{subfigure}[t]{0.32\textwidth}
        \centering
        \includegraphics[width=1.05\linewidth]{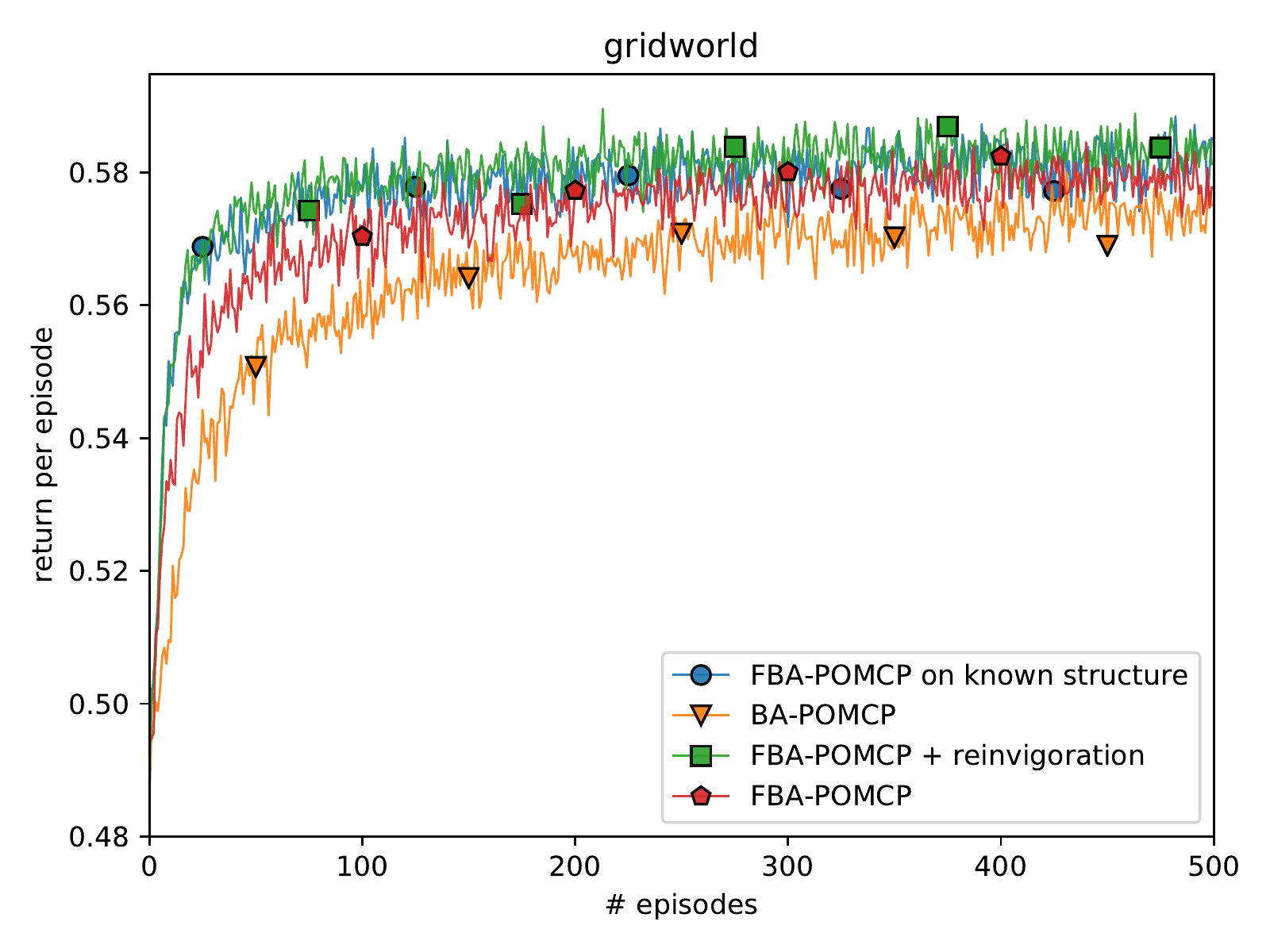}
    \end{subfigure}
    ~ 
    \begin{subfigure}[t]{0.32\textwidth}
        \centering
        \includegraphics[width=1.05\linewidth]{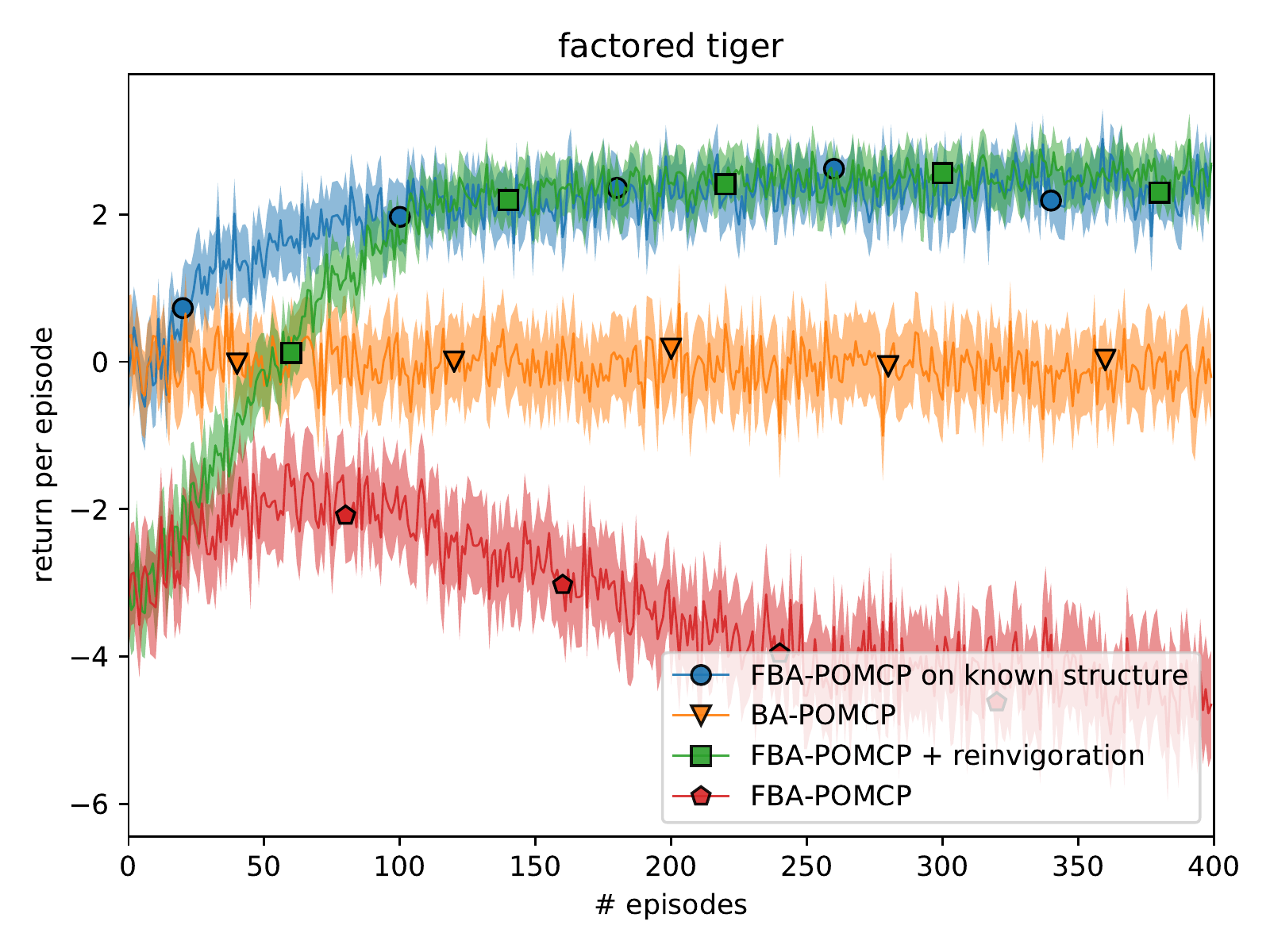}
    \end{subfigure}
    \captionsetup{justification=centering}
    \caption{Average return on the \ca\ (left), \gw\ (middle) and \ftiger\ (right) problem. \\ The shaded areas indicate the $95\%$ confidence interval.}
    \label{fig:results}
\end{figure*}

\subsection{Analysis} 

We compare {\color{green} \ourmethod} to $3$ other solution methods
(\cref{fig:results}). We consider the {\color{orange}\bapomcpAgent} agent,
as the baseline approach that ignores factorization and attempts to learn
in the problems framed as \bapomdp s. A second approach called
{\color{blue}\cheatAgent} acts with complete prior knowledge of the
structure of the dynamics and could be considered as a best-case scenario.
This method requires additional knowledge of the domain and thus can be
considered as `cheating' compared to the other methods\footnote
{
    Note, however that it is more common to know whether domain features
    share dependencies than it is to know the true probabilities
}.
Thirdly we test an agent {\color{red}\noreinvigAgent} with the same prior
knowledge as \ourmethod, except that it does not reinvigorate its belief.
The comparison with this approach highlights the contribution of the
reinvigoration step proposed by us to keep a healthy distribution over the
structure of the dynamics. \\

In order to produce statistically significant results we ran the
experiments described above a large number of times. In these experiments,
the parameters of the \mcts\ planning and \is\ belief update were equal
across all solution methods. We refer to the appendix for more details. 
We first make the general observation that \ourmethod\ consistently either
outperforms or is on par with the other methods, even compared to the
\cheatAgent\ approach whose prior knowledge is more accurate. \\

In the \ca\ domain \ourmethod\ outperforms even the agent with complete
knowledge to the model topology with statistical significance \cheatAgent.
This implies that particle reinvigoration (with respect to the true,
complete posterior) is beneficial even when structure degeneracy is not a
problem, because it improves the belief over the model by better
approximating the distribution over $\chi$. The dynamics of this problem
are particularly subtle, causing the belief over the graph topologies in
the \noreinvigAgent\ agent to converge to something different from the true
model. As a result, it is unable to exploit the compactness of the true
underlying dynamics and its performance is similar to \bapomcpAgent. None
of the agents in the \ca\ problem (left graph) have converged yet due to
the lack of learning time provided in the $500$ episodes (which we cut off
in the interest of time). \\

\gw\ has comparatively less subtle transitions, and all methods show a
generally quicker learning pace compared to other domains (centre figure).
Nevertheless, \bapomcpAgent\ has not converged to the true model yet after
$500$ episodes whereas the factored approaches (in particular \ourmethod\
and \cheatAgent) do so after less than $200$. \\

The results on the \ftiger\ problem (right figure) show significantly
different behaviour. Firstly, the initial performance of both \ourmethod\
and \noreinvigAgent\ is worse than the \bapomcpAgent\ and \cheatAgent. The
reason becomes obvious once you realize that due to the uniform prior over
the structure, half of the models in the initial belief contain topologies
that cannot represent the intended prior counts. This leads to a change in
the initial belief and thus the initial performance is different. 
With a uniform prior over the structure and without reinvigoration, the
agent could accidentally end up converging to a model structure that is
unable to express the true underlying dynamics. This is shown spectacularly
by \noreinvigAgent, where the average performance \emph{reduces} over time.
More detailed qualitative analysis showed that in most runs the agent
actually performs similar to the other \fbapomdp\ agents, however, every
now and then the belief converges to a model structure that does not
include the tiger location as parent feature in the observation model,
leading to a policy that opens doors randomly with an expected return of
$-45$. 
Lastly, the lack of improvement of the \bapomcpAgent\ agent emphasises the
need of factored representations most of all. Due to the large state space
the number of variables in the observation model grows too large to learn
individually. As a result, even $400$ episodes are not enough for the agent
to learn a model.

    \section{Related work}
Much of the recent work in \rl\ in partially observable environments has
been in applications of \drl\ to \pomdp s. To tackle the issue of
remembering past observations, researchers have attempted to employ
Recurrent networks~\cite{hausknecht2015deep,zhu2018improving}. Others have
introduced inductive biases into the network in order to learn a generative
model to imitate belief updates~\cite{igl2018deep}. While these approaches
are able to tackle large-scale problems, they are not Bayesian and hence do
not share the same theoretical guarantees. \\

More traditional approaches include the U-Tree
algorithm~\cite{mccallum1996reinforcement} (and its modifications),
EM-based algorithms such as~\cite{liu2013online} and policy gradient
descent methods~\cite{baxter2000direct}. One of their main drawbacks is
that they do not address the fundamental challenge of the
exploration-exploitation trade-off in \pomdp s. \\

There are other approaches that directly try to address this issue. The
Infinite-RPR~\cite{liu2011infinite} is an example of a model-free approach.
The Infinite-\pomdp~\cite{doshi2015bayesian} is an example of an
model-based solution method that attempts to do it in a model-based fashion
as well. Their approach is similar in the sense that they learn a model in
a Bayesian matter, however their assumptions of prior knowledge and about
what is being learned are different.

    \section{Conclusion}
This paper addresses the void for \brl\ models and methods at the
intersection of factored models and partially observable domains.
Our approach describes the dynamics of the \pomdp\ in terms of graphical
models and allows the agent to maintain a joint belief over the state, and
both the graph structure and \cpt\ parameters simultaneously. Alongside the
framework we introduced \fbapomcp, a solution method, which consists of an
extension of \mcts\ to \fbapomdp s, in addition to a particle
reinvigorating belief tracking algorithm.
The method is guaranteed to converge to the optimal policy with respect to
the initial, as both the planner and the belief update are unbiased. Lastly
we compared it to the current state-of-the-art approach on the $3$
different domains.
The results show the significance of representing and recognizing
independent features, as our method either outperforms \bapomdp\ based
agents or is able to learn in scenarios where tabular methods are not
feasible at all.

    \section*{Acknowledgements}
    Christopher Amato and Sammie Katt are funded by NSF Grant \#1734497, Frans
Oliehoek is funded by EPSRC First Grant EP/R001227/1, and ERC Starting Grant
\#758824---INFLUENCE.

    \bibliographystyle{acm}
    \bibliography{ref}  

\end{document}